\documentclass[runningheads]{llncs}
\usepackage{booktabs}
\usepackage[T1]{fontenc}
\usepackage{graphicx}
\usepackage{url}
\makeatletter
\g@addto@macro{\UrlBreaks}{\UrlOrds}
\makeatother
\usepackage{subcaption}
\usepackage{listings}       
\usepackage{xcolor}
\definecolor{backcolour}{RGB}{246, 246, 246}   
\definecolor{codegreen}{RGB}{16, 124, 2}       
\definecolor{codepurple}{RGB}{170, 0, 217}     
\definecolor{codered}{RGB}{154, 0, 18}         
\lstdefinestyle{gcolabstyle}{
  basicstyle=\ttfamily\small,
  backgroundcolor=\color{backcolour},   
  commentstyle=\itshape\color{codegreen},
  keywordstyle=\color{codepurple},
  stringstyle=\color{codered},
  numberstyle=\ttfamily\footnotesize\color{darkgray}, 
  breakatwhitespace=false,         
  breaklines=true,                 
  captionpos=b,                    
  keepspaces=true,                 
  numbers=left,                    
  numbersep=5pt,                  
  showspaces=false,                
  showstringspaces=false,
  showtabs=false,                  
  tabsize=2
}
\usepackage{makecell} 
\begin{document}
\title{Nobody Truly Agrees on Sentiment: Humans, Bespoke Tools, and LLMs Struggle with Social Media Texts}
\titlerunning{Nobody Truly Agrees on Sentiment}
%
\author{Himarsha R. Jayanetti\inst{1}\orcidID{0000-1111-2222-3333} \and 
Sivakanesan Dhanushkanda\inst{1}\orcidID{0000-0001-6452-5576} \and
Shuai Hao\inst{1}\orcidID{0000-0001-7483-5252} \and
Michael L. Nelson\inst{1}\orcidID{0000-0003-3749-8116} \and
Michele C. Weigle\inst{1}\orcidID{0000-0002-2787-7166}
}

\authorrunning{H. R. Jayanetti et al.}
\institute{Old Dominion University, Norfolk, Virginia, USA \\
\email{\{hjaya002,sdhan005,shao\}@odu.edu}
\email{\{mln,mweigle\}@cs.odu.edu}
}

\maketitle              
\begin{abstract}
Social media is a rich source of real-time public sentiment, but widely used sentiment analysis tools are often applied without understanding their limitations. In this study, we evaluate the inter-rater reliability of three bespoke sentiment analysis tools (TextBlob, VADER, and Twitter-roBERTa-base) and three large language models (LLMs: Qwen3-32B, GPT-OSS-120B, Llama-4-Maverick-17B) against six human raters across 100 tweets. We measured agreement using two statistical measures: Cohen’s kappa for pairwise comparisons and Fleiss’ kappa for multiple raters. Even among the human raters, our results showed only fair agreement, highlighting the subjectivity of sentiment analysis. Higher agreement was observed under the binary sentiment classification (negative vs. non-negative and positive vs. non-positive) than under the three-class classification across both humans and automated tools. The Twitter-roBERTa-base model showed the strongest alignment with human ratings, outperforming both bespoke sentiment tools and LLMs, particularly in distinguishing negative versus non-negative sentiment. LLMs showed substantial agreement among themselves and moderate to substantial alignment with humans, performing better in positive vs. non-positive classifications. Our findings underscore that domain-specific fine-tuning remains crucial for reliable social media sentiment analysis, and human-centered evaluation remains essential for establishing gold-standard labels.

\end{abstract}
\section{Introduction}
    In recent years, social media has become a valuable tool to capture real-time public sentiment. Sentiment analysis, a key task in natural language processing (NLP), enables the classification of such text as positive, negative, or neutral. However, the accessibility of existing bespoke sentiment analysis tools has led to their use without fully understanding their limitations. Running text through a tool and accepting its output at face value can be misleading, especially when the nuances of human interpretation are overlooked. Through our study we learned that even humans do not consistently agree on the sentiment of short, context-poor posts, and automated tools inherit the same ambiguity.
    
    In this study, we provide an empirical examination of sentiment analysis reliability by comparing both bespoke tools and modern Large Language Models (LLMs) directly against human judgment. Specifically, we analyzed the sentiment of 100 randomly selected tweets, small enough for detailed, qualitative review using three bespoke sentiment analysis tools (TextBlob~\cite{loria2018textblob}, VADER~\cite{hutto2014vader}, and Twitter-roBERTa-base model~\cite{barbieri-etal-2020-tweeteval}) as well as three large language models (Qwen3-32B~\cite{qwen3technicalreport,qwen_hf}, GPT-OSS-120B~\cite{gpt-oss_hf}, and Llama-4-Maverick-17B~\cite{llama_hf}), and compared their outputs with the ratings of six human raters. The deliberately small dataset enabled detailed qualitative review and independent manual annotation by all six raters. To measure agreement, we calculated Cohen’s kappa~\cite{cohen_kappa} for each pair of raters (including both humans and tools) and Fleiss’ kappa~\cite{fleiss_kappa} for multiple raters. 
    
    The main contribution of this work is a direct comparison of three sentiment analysis tools and three LLMs with multiple human raters, using quantitative measures of inter-rater agreement. Our results revealed a cautionary truth: no tool, LLM or otherwise, reliably matches human sentiment judgments, and even human raters only show fair agreement with each other. To address this, we took a majority vote among human raters and then compared it with the tools. Notably, the Twitter-roBERTa-base model showed the best alignment with human rating. We also compared three-class (positive, negative, neutral) and binary (negative vs. non-negative, positive vs. non-positive) sentiment classification. Our findings show higher agreement under the binary sentiment classification than under the three-class classification, with negative vs. non-negative distinctions achieving the strongest agreement. Our study provides practical insights for researchers and practitioners on selecting appropriate methods for analyzing social media sentiment, while highlighting that both bespoke tools and modern LLMs struggle to fully capture the nuances of human judgment in short-form social media texts. The persistence of this challenge, along with the assumption that newer or larger models yield more reliable results, underscores the need for careful evaluation and responsible use of automated methods. Sentiment analysis of short social media posts remains inherently noisy, requiring researchers to approach automated tools, old and new, with appropriate caution.






\section{Background and Related Work}
Sentiment analysis~\cite{1250949,liu2010sentiment} is a branch of NLP that examines people’s expressions of emotions, opinions, or stances on a topic. It considers both polarity (positive, negative, or neutral) and valence (emotional intensity) to classify text~\cite{tian2018polarity}. Sentiment analysis is used in many areas including marketing, finance, tourism, politics, and many other fields. Sentiment analysis is also widely applied to social media text, particularly Twitter (now X), where posts are brief, informal, and often filled with slang, hashtags, and sarcasm~\cite{wang2022sentiment,albladi2025sentiment,raisa2021review}. 



For this study, we focused on three representative sentiment analysis tools as well as three LLMs:
\begin{enumerate}
\item TextBlob: A Python library for processing textual data that includes a sentiment analysis feature that returns a Sentiment (polarity, subjectivity) tuple, where polarity ranges from [-1.0, 1.0] and subjectivity ranges from [0.0, 1.0]. We only used the polarity score to assign sentiment labels (positive, neutral, and negative) in this study.


\item VADER: A rule-based sentiment analysis model specifically developed for social media text. 


\item Twitter-roBERTa-base model: A model trained on 58M tweets and fine-tuned for sentiment analysis with the TweetEval benchmark in 2020~\cite{rosenthal2017semeval,2022tweetnlp}. 

\item LLMs: Qwen3-32B, GPT-OSS-120B, and Llama-4-Maverick-17B were tested with zero- and one-shot prompting. One-shot prompting performed slightly better or comparable to zero-shot. Since zero-shot results did not affect the overall conclusions, we report only one-shot results for clarity. The results for the zero-shot experiments, along with the prompts used for both the zero-shot and one-shot settings, are available in the project GitHub repository~\cite{dataset_100tweets}.



\end{enumerate}


Several studies have used TextBlob, VADER, and BERT-based models for analyzing public sentiment on social media platforms~\cite{aljedaani2022sentiment,marrapu2025sentiment,ccilgin2022twitter,bello2023bert}. Comparative studies show that VADER often outperforms TextBlob for short, casual posts, while transformer-based models excel in capturing nuanced, context-dependent sentiment~\cite{dahal_gupta_budhathoki_2025,ccoya2023,bello2023bert}. Despite these advances, TextBlob and VADER remain widely used due to their simplicity and accessibility~\cite{doan2025sentiment,irfan2025sentiment}. More recently, transformer-based models, like BERT~\cite{devlin2019bert} and RoBERTa~\cite{liu2019roberta}, have achieved state-of-the-art performance by capturing complex linguistic and contextual dependencies, particularly when fine-tuned to a specific domain~\cite{nguyen2020bertweet,semary2023improving,prytula2024fine,kumar2024fusion}. Our study offers a comparison of TextBlob and VADER with the fine-tuned Twitter-roBERTa-base model and human judgments, highlighting their strengths and limitations. In another recent study, Kwon et al.~\cite{KWON2024114570} used LLMs to analyze public sentiment on Twitter. They labeled each tweet using seven libraries, including TextBlob and VADER and applied a majority voting approach to label a dataset of 1.26 million tweets. Their evaluation showed that LLMs such as BERT, GPT-2, and Llama-2 outperformed traditional machine learning classifiers.

In contrast to those previously mentioned works, our study takes a fine-grained approach by evaluating 100 tweets with three sentiment analysis tools and six independent human raters, rather than relying on pre-labeled datasets or large-scale automated labeling. This design enabled us to directly compare output from tools to human judgments and also capture the diversity and subjectivity inherent in human judgment, particularly for informal social media content. By including multiple raters, we can quantify inter-rater agreement and evaluate how closely each tool aligns with human perception, offering insights into tool strengths and limitations in practical social media applications.

\section{Methodology}

Our dataset consists of a subset of 100 tweets~\cite{dataset_100tweets} containing the keyword ``Site C, Khayelitsha'' posted between January 2022 to December 2024, collected as part of a study to assess residents' perceptions of safety and security in Khayelitsha Township, South Africa. We then generated the sentiment labels by running the three sentiment analysis tools and three LLMs and also by gathering data from six human raters. Those human raters were Computer Science graduate students from our lab with substantial exposure to social media content and sentiment analysis tasks. Hereafter, the six human raters are referred to as R1–R6; the three sentiment analysis tools are referred to as TB (TextBlob), VD (VADER), and RB (Twitter-RoBERTa-base); and the three LLMs are referred to as QW (Qwen3-32B), GP (GPT-OSS-120B), and LM (Llama-4-Maverick-17B). 

To evaluate the agreement between sentiment analysis tools and human perception, we followed four steps. First, we measured pairwise agreement between human raters and tools using Cohen’s kappa, which provides insight into consistency at the individual rater level. We also tested if simplifying sentiment into binary categories (e.g., negative vs. non-negative and positive vs. non-positive) improves agreement. Next, we extended the analysis to include multiple raters simultaneously by applying Fleiss’ kappa. As we noticed that the human ratings also varied, we used a majority voting label between human raters which served as a label for evaluating tool performance. No ties occurred during the majority voting process; therefore, no tie-breaking procedure was required. 

\subsection{Measuring Inter-Rater Agreement: Cohen’s Kappa}
Cohen's kappa coefficient is a statistic that is used to measure inter-rater agreement between two raters for categorical items. The coefficient ranges from -1 to 1, with 1 indicating perfect agreement, 0 if there is no agreement other than what would be expected by chance, and negative values indicating disagreement (agreement worse than chance). Kappa values can be interpreted in the following ranges: -1.00-0.00 (poor agreement), 0.01-0.20 (slight), 0.21-0.40 (fair), 0.41-0.60 (moderate), 0.61-0.80 (substantial), and 0.81-1.00 (almost perfect).

\begin{figure*}[!ht]
    \centering
    \includegraphics[width=\textwidth]{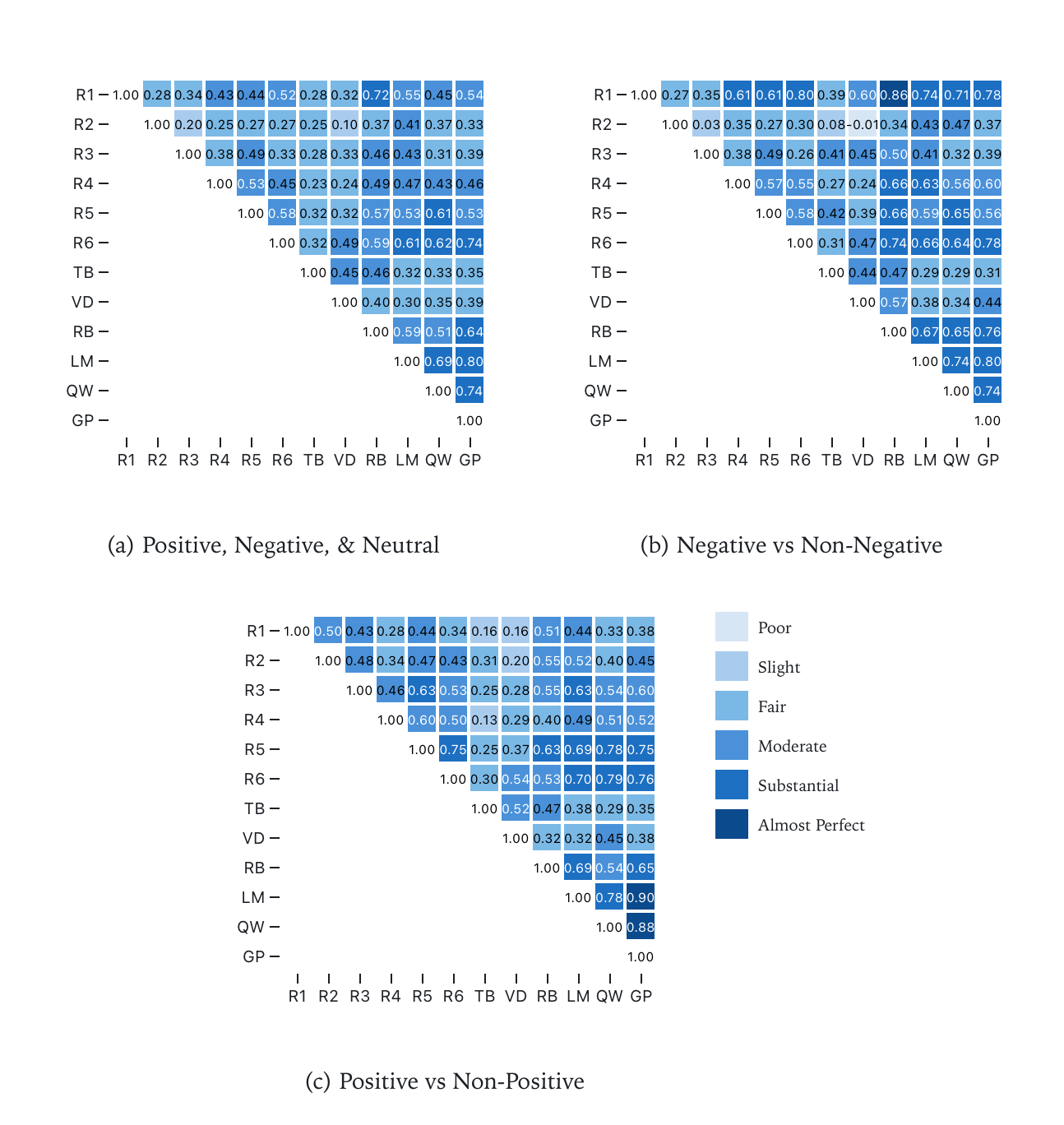} 
    \caption{Cohen's Kappa (a) Positive, Negative, and Neutral, (b) Negative vs Non-Negative, (c) Positive vs Non-Positive. Darker values indicate better agreement.}
    \label{fig:cohen_kappa}
\end{figure*}


We hypothesized that classifying sentiment as negative vs. non-negative (or even positive vs. non-positive) might yield higher agreement than the traditional positive/negative/neutral categories. This is because humans are generally more adept at identifying distinct positive or negative sentiments, while neutral statements tend to be more ambiguous and prone to subjective interpretation. To test our theory, we calculated the Cohen’s kappa for negative vs. non-negative (Fig.~\ref{fig:cohen_kappa}b) and positive vs. non-positive (Fig.~\ref{fig:cohen_kappa}c).

\subsection{Multi-Rater Agreement: Fleiss’ Kappa}

Fleiss's kappa is a generalization of Cohen's kappa for more than two raters. In Table~\ref{tab:fliess_kappa_combined_rounded}, we calculated Fleiss' kappa to assess inter-rater agreement across multiple raters and tools for various sentiment classification tasks: positive/negative/neutral, negative/non-negative, and positive/non-positive.

\section{Results}

We first calculated the Cohen’s kappa for each pair of raters (Fig.~\ref{fig:cohen_kappa}a). The kappa values between human raters and tools vary, with some raters showing moderate agreement (e.g., R1 and R6 with a kappa of 0.52) and others showing fair agreement (e.g., R1 and R2 with a kappa of 0.28). The sentiment analysis tools (TB, VD, and RB) generally have slight to fair agreement with the human raters (R1-R6), indicating that the tools are less consistent with human raters. Among those tools, RB has the highest agreement with a human rater (kappa value of 0.72 with R1). In contrast, the LLMs (LM, GP, and QW) have higher agreement with human raters than other sentiment analysis tools, generally achieving fair to moderate agreement with human raters (R1-R6). Among LLMs, GP has the highest agreement with a human rater (kappa value of 0.74 with R6). Additionally, agreement among the LLMs themselves was substantial, with kappa value of 0.69 (LM and QW), 0.74 (GP and QW), and 0.80 (LM and GP).


\subsection{Higher Inter-Rater Agreement Under Binary Sentiment Classification}
The kappa values between raters for negative vs. non-negative show significantly higher agreement compared to the three-class sentiment classification in Fig.~\ref{fig:cohen_kappa}a. For example, in Fig.~\ref{fig:cohen_kappa}b, human raters R1 and R6 achieve almost perfect agreement (kappa = 0.80). RB exhibits the highest agreement with the raters (for example, RB with R1 having an almost perfect kappa value of 0.86), which is similar to what we saw in the three-class classification, but with noticeably stronger agreement in the binary negative vs. non-negative classification. Similarly, the LLMs show strong alignment with human judgments in this binary setting: GP achieves substantial agreement with R1 and R6 (kappa = 0.78), LM with R1 (kappa = 0.74), and QW with R1 (kappa = 0.65). TB and VD still show slight and fair agreement. In particular, VD exhibits poor agreement with R2 (kappa = -0.01); however, this low value is consistent with R2’s generally weak agreement across multiple tools and raters.


As shown in Fig.~\ref{fig:cohen_kappa}c, the kappa values for the positive vs. non-positive classification also show a moderate increase in agreement compared to the three-class classification in Fig.~\ref{fig:cohen_kappa}a, but the values are generally lower than the negative vs. non-negative classification in Fig.~\ref{fig:cohen_kappa}b. For example, R1 and R6 have slight agreement with a kappa value of 0.34. However, as shown in Fig.~\ref{fig:cohen_kappa}b, they have almost perfect agreement (kappa value of 0.80) for the negative vs. non-negative classification. RB continues to show better performance in positive vs. non-positive binary classifications (with kappa values like 0.63 with R5). The LLMs showed higher agreement in positive vs. non-positive classification, with substantial to almost perfect agreement in some pairings. TB and VD show weaker agreement in the positive vs. non-positive classification compared to negative vs. non-negative.



In summary, higher agreement was observed under the binary sentiment classification compared to the three-class classification. For human raters and other non-LLM tools, agreement is generally stronger for negative vs. non-negative, suggesting that negative sentiment is more distinct and easier to identify in our dataset. However, the LLMs tend to agree more with humans on positive sentiment.

\subsection{Human Consensus via Majority Voting Improves Agreement with Automated Tools}


\begin{table}
    \centering\scriptsize
    \setlength{\tabcolsep}{3pt} 
    \caption{Inter-Rater Agreement (Fleiss’ Kappa). The highest kappa value for each classification is shown in \textbf{bold} and highlighted in \textcolor{red}{red}.}

    \begin{tabular}{lccc}
    \toprule
     & \makecell{Pos/Neg/Neutral} & \makecell{Neg/Non-Neg} & \makecell{Pos/Non-Pos} \\
    \midrule
    All Humans (R1-R6) & 0.39 & 0.45 & 0.48 \\
    All Tools (TB, VD, RB) & \textcolor{red}{\textbf{0.43}} & 0.49 & 0.43 \\
    Humans (R1-R6) with TB & 0.35 & 0.41 & 0.39 \\
    Humans (R1-R6) with VD & 0.35 & 0.42 & 0.42 \\
    Humans (R1-R6) with RB & \textcolor{red}{\textbf{0.43}} & \textcolor{red}{\textbf{0.51}} & \textcolor{red}{\textbf{0.50}} \\
    Humans (R1-R6) with TB, VD, RB & 0.38 & 0.45 & 0.40 \\
    \midrule
    All LLMs (one-shot) & \textcolor{red}{\textbf{0.74}} & \textcolor{red}{\textbf{0.76}} & \textcolor{red}{\textbf{0.86}} \\
    Humans with LLaMA (one-shot) & 0.42 & 0.49 & 0.51 \\
    Humans with Qwen (one-shot) & 0.41 & 0.49 & 0.51 \\
    Humans with GPT (one-shot) & 0.42 & 0.49 & 0.52 \\
    Humans with all LLMs (one-shot) & 0.46 & 0.54 & 0.57 \\
    \bottomrule
    \end{tabular}
    \label{tab:fliess_kappa_combined_rounded}
\end{table}


As shown in Table~\ref{tab:fliess_kappa_combined_rounded}, the individual human raters had varying interpretations of sentiment, leading to moderate inter-rater agreement at best among themselves and with the sentiment analysis tools. Agreement among LLMs was substantial to near-perfect, whereas agreement between LLMs and human raters remained at most moderate. To address this inconsistency and improve reliability, we applied a majority voting approach, aggregating the most common sentiment label among human raters (Table~\ref{tab:majority_voting_llms}). This allowed us to establish a more stable consensus, which we then used to compare against sentiment analysis tools and LLMs. We noticed that applying majority voting among human raters significantly improved consistency, making their collective judgment more aligned with automated tools.


\begin{table}
    \centering\scriptsize
    \setlength{\tabcolsep}{4pt} 
    \caption{Human Majority vs. Each Tool / LLM (Cohen’s Kappa). The highest kappa value for each classification is shown in \textbf{bold} and highlighted in \textcolor{red}{red}.}
    \begin{tabular}{lccc}
    \toprule
     & \makecell{Pos/Neg/Neutral} & \makecell{Neg/Non-Neg} & \makecell{Pos/Non-Pos} \\
    \midrule
    Majority vs.\ TB & 0.37 & 0.41 & 0.29 \\
    Majority vs.\ VD & 0.40 & 0.50 & 0.37 \\
    Majority vs.\ RB & \textcolor{red}{\textbf{0.73}} & \textcolor{red}{\textbf{0.88}} & \textcolor{red}{\textbf{0.61}} \\
    \midrule
    Majority vs.\ LLaMA (one-shot) & \textcolor{red}{\textbf{0.63}} & \textcolor{red}{\textbf{0.71}} & \textcolor{red}{\textbf{0.71}} \\
    Majority vs.\ Qwen (one-shot) & 0.48 & 0.60 & 0.61 \\
    Majority vs.\ GPT (one-shot) & 0.57 & 0.67 & 0.67 \\
    \bottomrule
    \end{tabular}
    \label{tab:majority_voting_llms}
\end{table}

RB shows almost perfect agreement with the majority vote of human raters, particularly in negative/non-negative (0.88). It also shows substantial agreement in positive/negative/neutral (0.73) and positive/non-positive (0.61). This indicates that RB is highly consistent with the human majority, especially when distinguishing between negative and non-negative sentiments, and also performs substantially in classifying positive/negative/neutral and positive/non-positive sentiment. TB and VD show fair/moderate agreement, with VD performing moderately in negative/non-negative (0.50) but still demonstrating lower kappa values compared to RB. TB performs the worst, especially in positive/non-positive (0.29), showing fair agreement with the majority vote of human raters. LLM-based models also exhibit strong alignment with the human majority vote. LM achieves substantial agreement and GP shows consistently moderate to substantial agreement. QW demonstrates moderate agreement overall, with comparatively lower performance in the multi-class positive/negative/neutral task. The agreement with the human majority is strongest for the binary sentiment classification for all LLMs. Despite the strong performance of LLMs, the domain-specific RB model remains the most aligned tool with the human majority overall, achieving the highest agreement except in the positive/non-positive setting where LLaMA attains the highest kappa value.

\section{Limitations and Future Work}
Although this study focused on 100 tweets, the deliberately curated dataset enabled independent annotation by six human raters and a detailed evaluation of inter-rater agreement. Expanding this approach to additional topics, geographic regions, and social media communities will help assess the broader applicability of these findings. While the evaluated sentiment analysis tools produce outputs of varying granularity, including continuous polarity scores (e.g., TextBlob and VADER) and class probabilities (e.g., RoBERTa), we converted all outputs into negative, neutral, and positive labels. This decision was made to ensure methodological consistency across all automated systems and human annotators, as the human reference annotations were collected as discrete sentiment categories rather than continuous ratings. We acknowledge that this simplification discards information about sentiment intensity and prediction uncertainty, which may provide additional insight into cases where models or annotators exhibit low confidence or where sentiment is inherently ambiguous. In addition, this study reports point estimates of agreement statistics without confidence intervals or formal significance testing; therefore, differences between agreement values should be interpreted descriptively rather than as statistically established. Future research could complement categorical agreement analyses with continuous or probability-based comparisons and report confidence intervals or hypothesis tests using larger datasets to better capture sentiment intensity, model uncertainty, and the statistical reliability of agreement estimates.

\section{Conclusions} 

We compared three sentiment analysis tools  (TextBlob, VADER, and Twitter-roBERTa-base model) and three LLMs (Qwen3-32B, GPT-OSS-120B, and Llama-4-Maverick-17B) on social media text against human judgments. 


\textbf{Higher agreement under binary sentiment classification:}
Higher agreement was observed under the negative vs. non-negative and positive vs. non-positive classification than under the three-class classification among human raters and between humans and automated tools. For sentiment tools and humans, negative vs. non-negative classification showed the strongest agreement, suggesting that negative sentiment is more distinct and easier to identify than positive sentiment. Interestingly, LLMs performed better on positive vs. non-positive classifications, achieving higher alignment with human judgments in detecting positive sentiment.

\textbf{Twitter-RoBERTa-base emerges as the most reliable tool in matching human sentiment classifications:} 
It consistently showed substantial/almost perfect agreement with the majority vote across all sentiment categories. This shows that carefully trained, domain-specific models can sometimes outperform even recent general-purpose LLMs.

\textbf{LLMs also show strong alignment with human sentiment:} LLMs exhibited substantial agreement among themselves and moderate to substantial agreement with the majority vote by humans. LLMs also performed better in binary sentiment classification, with positive vs. non-positive yielding relatively higher agreement. However, general-purpose LLMs, while powerful, do not automatically outperform classical sentiment analysis tools in social media text as they lack task-specific training on social media sentiment.

\textbf{TextBlob and VADER show weaker performance:}
Both TextBlob and VADER exhibit less agreement, particularly in positive/non-positive, indicating they might struggle with identifying positive sentiments as clearly as negative or non-negative ones. TextBlob showed the lowest agreement across all categories.

\textbf{Human-centered evaluation is important for reliable sentiment analysis:} Our results showed how even human raters still disagree, highlighting the difficulty of sentiment analysis of social media text. Multi-rater evaluation is essential for establishing a reliable gold standard. Although our dataset includes only 100 tweets, a well-curated, small-scale dataset allows for detailed analysis and multiple human ratings per tweet. Importantly, automated tools or LLMs should not be used to generate ground truth, as they may not always align well with human perception; human-verified labels remain the benchmark for evaluating sentiment.

Twitter-RoBERTa-base and the LLMs outperform classical lexicon-based tools for sentiment analysis because they apply transformer-based deep learning to understand the context of words. Twitter-RoBERTa-base, as a domain-specific model fine-tuned on TweetEval, is more effective at capturing sentiment than TextBlob and VADER, which rely on fixed rules and lexicons. In contrast, general-purpose LLMs are trained on large corpora but are not specifically adapted to social media text. While LLMs capture nuanced sentiment well, domain-specific fine-tuning could further improve their alignment with human judgments. As these tools continue to advance, our findings are proof positive that the future of sentiment analysis lies in models that truly understand the complexity of human emotions in text and that domain adaptation still remains crucial for reliable sentiment analysis. 

\begin{credits}
\subsubsection{\ackname} This project was supported in part through the Minerva Research Initiative, in partnership with the Air Force Office of Scientific Research (AFOSR) under grant number FA9550-22-1-0297. We would also like to thank David Calano, Lesley Frew, Kritika Garg, and Tarannum Zaki for their contributions to the evaluation dataset.
\end{credits}
%
%
%
%
\bibliographystyle{splncs04}
\bibliography{ref}
\end{document}